\documentclass[10pt,twocolumn,letterpaper]{article}

\usepackage[pagenumbers]{cvpr} 

\definecolor{cvprblue}{rgb}{0.21,0.49,0.74}
\usepackage[pagebackref,breaklinks,colorlinks,allcolors=cvprblue]{hyperref}
\usepackage{multirow}
\usepackage{array}
\usepackage{cellspace}
\usepackage{makecell}
\usepackage{xcolor}
\usepackage{colortbl}

\definecolor{lightblue}{rgb}{0.8,0.9,1}

\def\paperID{} 
\def\confName{CVPR}
\def\confYear{2026}

\title{Say Cheese! Detail-Preserving Portrait Collection Generation \\ via Natural Language Edits}

\author{Zelong Sun
,   Jiahui Wu 
,   Ying Ba
,   Dong Jing
,   Zhiwu Lu \thanks{Corresponding Author} \\
Gaoling School of Artificial Intelligence \\
Renmin University of China, Beijing, China\\
{\tt\small zelongsun@ruc.edu.cn, luzhiwu@ruc.edu.cn}}

\begin{document}

\twocolumn[{%
\renewcommand\twocolumn[1][]{#1}%
\maketitle

\vspace{-0.4in}
\begin{center}
\centering
\includegraphics[width=1.0\linewidth]{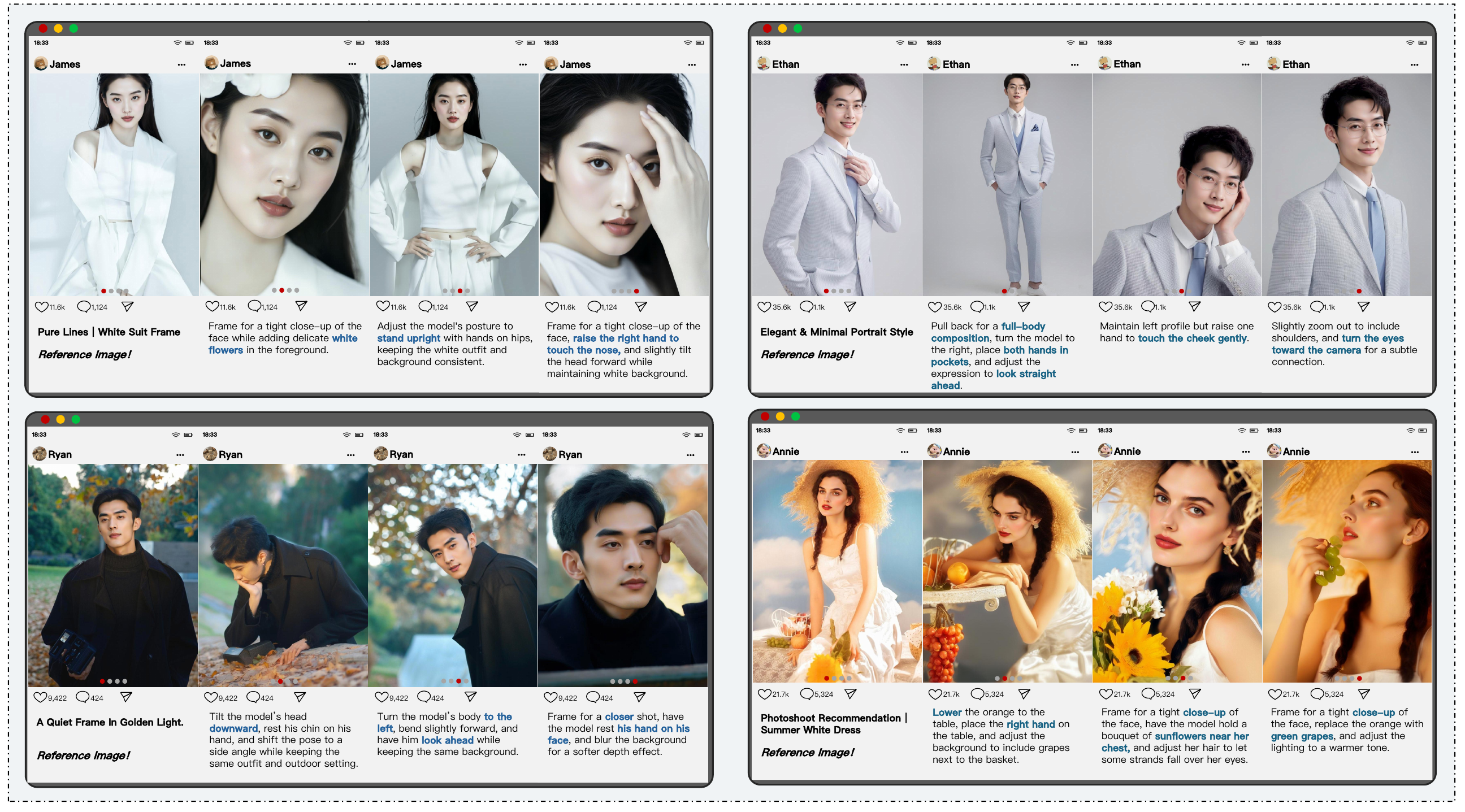} 
\vspace{-0.3in}
\captionof{figure}{\textbf{Generated Portrait Collection Generation (PCG) Examples.} Each group shows a reference image (left) and three generated results by our method with corresponding modification texts (below the photos). PCG introduces two key challenges: \textbf{(1) Complex modifications} requiring simultaneous changes in pose, camera viewpoint, and spatial layout; \textbf{(2) Detail preservation} maintaining fine-grained appearance characteristics, such as makeup, clothingand accessories.}
\label{fig_intro}
\end{center}%
}]

\begin{abstract}
As social media platforms proliferate, users increasingly demand intuitive ways to create diverse, high-quality portrait collections. In this work, we introduce Portrait Collection Generation (PCG), a novel task that generates coherent portrait collections by editing a reference portrait image through natural language instructions. This task poses two unique challenges to existing methods: (1) \textbf{complex multi-attribute modifications} such as pose, spatial layout, and camera viewpoint; and (2) \textbf{high-fidelity detail preservation} including identity, clothing, and accessories. To address these challenges, we propose \textbf{CHEESE}, the first large-scale PCG dataset containing 24K portrait collections and 573K samples with high-quality modification text annotations, constructed through an Large Vison-Language Model-based pipeline with inversion-based verification. We further propose \textbf{SCheese}, a framework that combines text-guided generation with hierarchical identity and detail preservation. SCheese employs adaptive feature fusion mechanism to maintain identity consistency, and ConsistencyNet to inject fine-grained features for detail consistency. Comprehensive experiments validate the effectiveness of CHEESE in advancing PCG, with SCheese achieving state-of-the-art performance.

\end{abstract}

\vspace{-0.2in}

\section{Introduction}
\label{sec:intro}

\begin{figure}[t!]
\centering  
\includegraphics[width=0.97\linewidth]{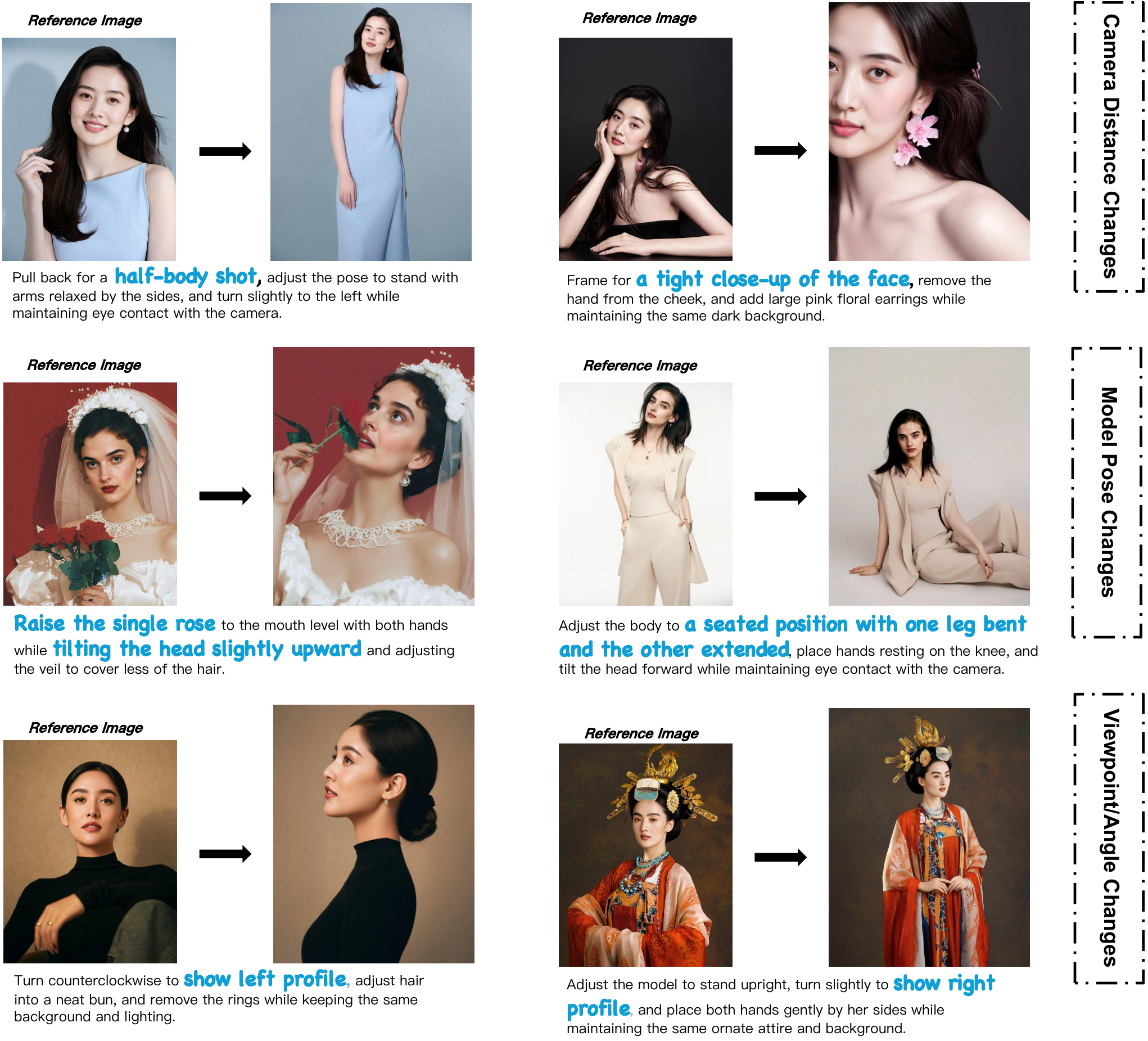} 
\vspace{-0.08in}
\caption{\textbf{Generated Examples of Several Editing Challenges in PCG:} camera distance, pose variation, and viewpoint transformation while preserving fine-grained details. 
}
\label{example}
\vspace{-0.3in}
\end{figure}

In the digital era, portrait collections has become an essential way for people to document personal moments and express their identities, as shown in Figure~\ref{fig_intro}. With the advancement of generative AI~\cite{ho2020denoising,rombach2022high,imagen, ramesh2022hierarchical}, users increasingly demand intuitive ways to create diverse and high-quality portrait collections. To address this demand, we propose Portrait Collection Generation (PCG), a novel task that enables users to create portrait collections using a high-quality reference image and natural language instructions. Compared to existing image editing tasks, PCG introduces two key challenges, as illustrated in Figure~\ref{example}: (1) \textbf{complex modification intents} spanning camera distance, pose, and viewpoint; and (2) \textbf{high-fidelity detail preservation} from the reference, including makeup, clothing, accessories, and fine identity cues, consistently across the collection.

Regarding the first challenge, ControlNet~\cite{contronet} and ControlNet++~\cite{li2024controlnet++} achieve fine-grained structure control by incorporating conditioning signals such as depth maps and Canny edges. However, these conditions often constrain spatial layout diversity, limiting the flexibility required for portrait photography. Other methods~\cite{brooks2023instructpix2pix, zhang2023magicbrush}, perform editing through text instructions by fine-tuning models on instruction-following datasets. However, their single-dimensional instructions often fail to meet users' increasingly complex editing requirements. As shown in Figure~\ref{fig_intro} and~\ref{example}, PCG frequently involve modifications in spatial layout, model pose, and viewpoint that are increasingly necessary to create diverse, expressive portrait collections.

Regarding the second challenge, existing fine-tuning methods such as DreamBooth~\cite{ruiz2023dreambooth} and LoRA~\cite{hu2022lora} can improve consistency with reference images using a few reference samples. However, these methods require per-subject training, limiting their scalability in practical applications. Zero-shot alternatives like IP-Adapter~\cite{ye2023ip} and InstantID~\cite{wang2024instantid} offer faster solutions but fall short in maintaining fine-grained details under complex modifications. Their fundamental limitation lies in their reliance on high-level semantic embeddings, which fail to preserve pixel-level details. PCG presents a more demanding challenge: simultaneously preserving both identity consistency and fine-grained detail consistency, illustrated in Figure~\ref{fig_intro} and~\ref{example}.

To address these challenges, we first propose \textbf{CHEESE}, the first large-scale dataset specifically designed for PCG, emphasizing both complex multi-attribute modifications and high-fidelity detail preservation. We collect 24K high-resolution portrait collections from the web, encompassing diverse scenarios (e.g., indoor and outdoor), styles (e.g., ancient, ethnic, and modern), and variations (e.g., pose changes, camera angles/distance, and spatial layouts). 
We then pair images from the same portrait collection and employ a Large Vision-Language Model (LVLM) to obtain detailed, compositionally rich modification texts, resulting in 575K triplets. Note that modification texts can range from single-dimensional changes to complex combinations of multiple transformations.
To ensure annotation quality, we introduce an inversion-based verification mechanism that provides quantitative scores for annotations and regenerates low-scoring ones. By sampling real-world, professional portrait collections, CHEESE ensures the \textbf{high-quality}, \textbf{diversity}, \textbf{creativity}, \textbf{content richness}, and \textbf{high-fidelity detail consistency} required for PCG.


We further propose \textbf{SCheese}, a framework that combines text-guided generation capabilities with identity- and detail-preserving mechanisms. For complex modification intents, we adopt a text-conditioned generation backbone that processes modification text to guide compositional transformations. For identity consistency, unlike IP-Adapter that only encodes the reference image, we fuse modification text and reference image features to directly obtain features that approximate the target image, supervised by an additional alignment loss. Such fused features provide more precise conditioning signals that align with the desired target characteristics, reducing ambiguity and improving instruction-following capability. During training, we apply teacher forcing by probabilistically replacing the fused features with target image features to provide direct supervision. For detail preservation, we propose ConsistencyNet, which extracts multi-scale fine-grained features from the reference image and injects them throughout the generation process via a Decoupled-Attention mechanism. Together, as shown in Figure~\ref{example}, SCheese can balance complex modifications with fine-grained detail preservation.

Our main contributions are threefold: (1) We propose \textbf{CHEESE}, the first large-scale dataset for PCG featuring complex modification annotations and detail consistency; (2) We design \textbf{SCheese}, a novel framework to balance complex modifications with fine-grained detail preservation; (3) Comprehensive experiments validate the effectiveness of CHEESE in advancing PCG, with SCheese achieving state-of-the-art performance in handling complex edits with identity and fine-grained details consistency.

\section{Related Work}

\textbf{Text-Guided and Conditioned Image Editing.} Text-guided image editing enables semantic modifications via natural language. Early diffusion-based approaches~\cite{kim2022diffusionclip,imagen} demonstrated text conditioning for generation. Recent works~\cite{brooks2023instructpix2pix,zhang2023magicbrush} fine-tuned diffusion models on instruction-following datasets. However, these methods primarily handle simple edits and struggle with complex, multi-faceted instructions requiring simultaneous changes across pose, layout, and camera parameters~\cite{ho2020denoising,rombach2022high}. 
Beyond text-only control, methods like ControlNet~\cite{contronet} and T2I-Adapter~\cite{T2i-adapter} introduce spatial conditioning via sketches, depth maps, and segmentation masks. ControlNet++~\cite{li2024controlnet++} and Uni-ControlNet~\cite{Uni-controlnet} further enhance spatial control integration. While these methods offer flexibility, they focus on structural conditioning or high-level semantics and lack capabilities for fine-grained identity preservation, which is critical for PCG task. Moreover, they typically operate on single-image edits without addressing consistency requirements for portrait collection generation.

\textbf{Personalization in Diffusion Models.} 
Image personalization approaches can be categorized into tuning-free methods~\cite{he2024imagine, jiang2024comat, li2024photomaker, qi2024deadiff, wang2024instantstyle, wang2024instantstyle, wang2024instantid, zhang2023adding}  and tuning-based methods~\cite{gal2022image, hu2022lora, ruiz2023dreambooth}. DreamBooth~\cite{ruiz2023dreambooth} fine-tunes diffusion models on reference images to learn subject-specific representations, while LoRA~\cite{hu2022lora} offers parameter-efficient fine-tuning via low-rank matrices. However, both require per-subject training and multiple references, making them computationally intensive. IP-Adapter~\cite{ye2023ip} enables zero-shot image-conditioned generation via lightweight adapters, and InstantID~\cite{wang2024instantid} improves facial identity preservation using face recognition features. IPAdapter-Instruct~\cite{rowles2024ipadapter_instruct} combines image prompts with instructions but relies on carefully crafted prompts. Alternative approaches~\cite{chen2023subject,huang2024group,ma2024subject} depend on manually curated datasets or are limited to simple attribute copying. These limitations become evident in PCG, where methods must simultaneously accommodate multi-attribute modifications while maintaining fine-grained detail preservation across collections.

\begin{figure}[t!]
\centering  
\includegraphics[width=0.99\linewidth]{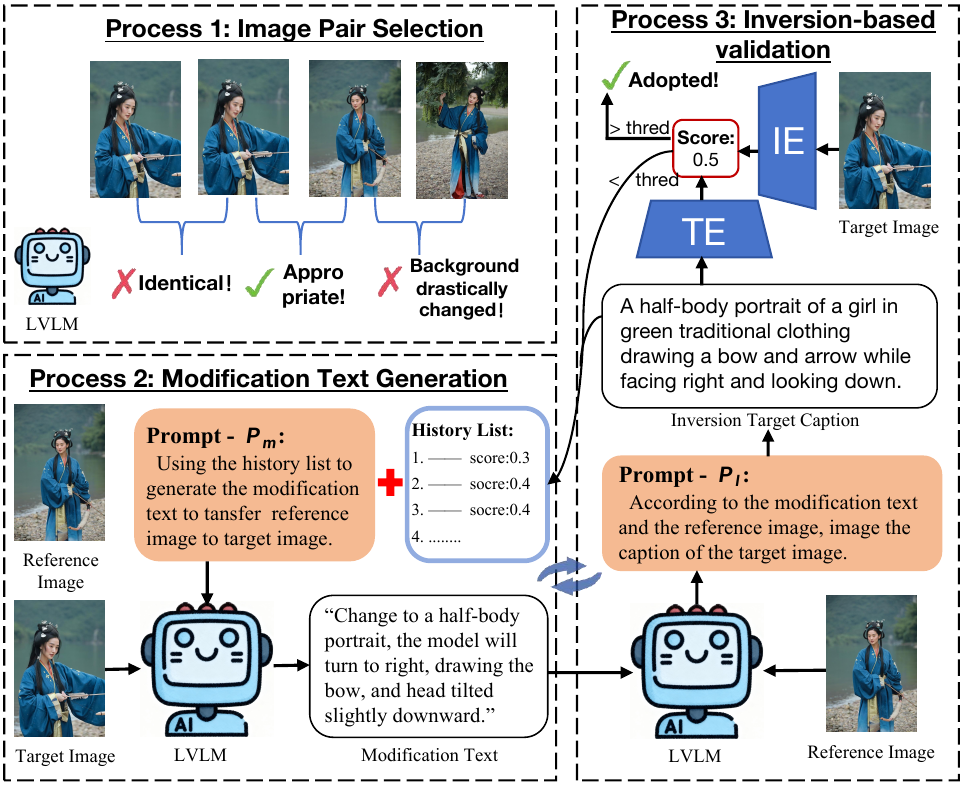} 
\vspace{-0.1in}
\caption{\textbf{The Dataset Construction Pipeline of CHEESE.} (1) We first use the LVLM filters out near-duplicate pairs and pairs with excessive background changes. (2) For each filtered pair, an LVLM generates a natural-language modification text describing the transformation. (3) We then use the LVLM to inversion the target caption and compute CLIP score with the target image. We then iteratively refine the annotation if the score falls below a threshold. $IE$, $TE$ denote the image and text encoder of CLIP.}
\label{data_pipeline}  
\vspace{-0.15in}
\end{figure}

\begin{figure*}[t!]
\centering  
\includegraphics[width=0.98\linewidth]{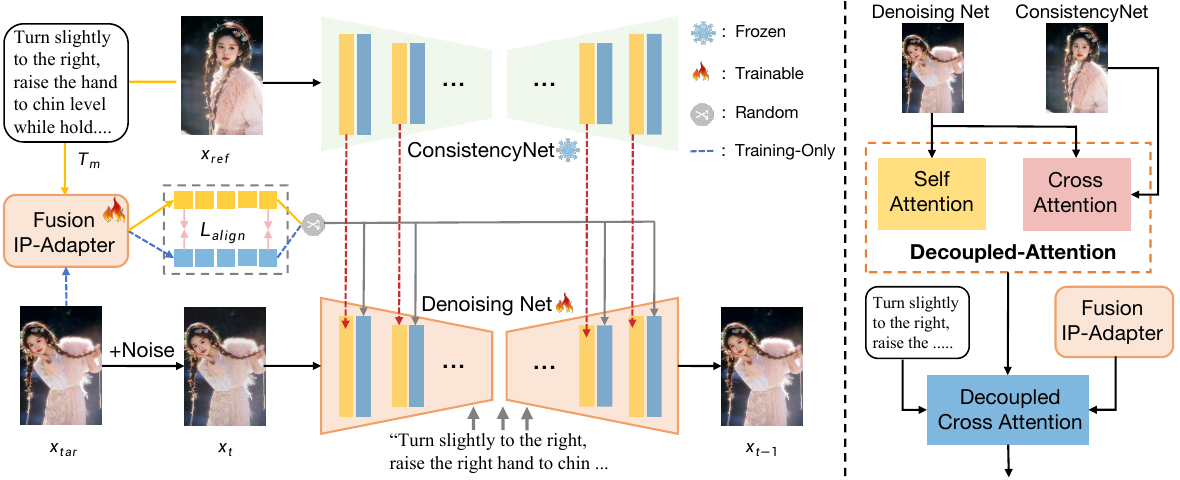} 
\vspace{-0.1in}
\caption{\textbf{Overview Architecture of SCheese:} 
\textbf{(Left)} Our model consists of (1) DenoisingNet which is a main UNet that processes target image, (2) Fusion IP-Adapter that fuses high-level semantics of reference image $I_r$ and modification text $T_m$ , and (3) ConsistencyNet that encodes low-level features of $I_r$. 
\textbf{(Right)} We propose a Decoupled-Attention mechanism, which consists of a self-attention module and a cross-attention module. The outputs of these two modules are averaged together and further fused with features from the text encoder and Fusion IP-Adapter through a decoupled cross-attention layer.}
\label{model_art}  
\vspace{-0.1in}
\end{figure*} 

\section{CHEESE}
In this section, we propose a large-scale dataset tailored for PCG, CHEESE $D=\{(I_r^{(n)},T_m^{(n)},I_t^{(n)})\}_{n=1}^N$, where $I_r$ and $I_t$ denotes the high-resolution reference and target image respectively, $T_m$ denotes the complex modification text, and $N$ denotes the triple number in $D$. 
The overview of our data construction pipeline in shown in Figure~\ref{data_pipeline}.

\noindent \textbf{Image Pair Selection.} We first collect portrait collections $A=\{A_k\}_{k=1}^K$ from the web, ensuring that each collection $A_k$ contains multiple high-resolution images of a consistent subject and style. For each collection $A_k$, we enumerate all within-collection pairs $(I_i, I_j)$ with $i \neq j$, and apply a LVLM to filter out near-duplicate pairs and those with excessive background or scene drift. This step removes trivial correspondences that would not support meaningful edits, while also discarding pairs where scene mismatch could confound identity preservation or detail fidelity. The result is a set of candidate pairs that balance visual diversity with subject and context coherence.

\noindent \textbf{Modification Text Annotation.} As shown in Figure~\ref{data_pipeline}, given a filtered pair $(I_r^{(n)},I_t^{(n)})$, the annotation stage prompts an LVLM to generate a modification text $T_m^{(n)}$ describing the transformation from $I_r$ to $I_t$. The prompt $P_{m}$ design explicitly encourages the model to capture changes in camera parameters (such as depth of field and viewpoint), spatial/layout adjustments, and subject-level variations including pose, expression, and orientation. This design promotes compositional coverage and helps disentangle multi-faceted edits that often co-occur in PCG. 
Notably, given varying degrees of differences between image pair, $T_m$ can range from single-dimensional changes to combinations of multiple transformations, aligning the diverse editing requirements in real-world.

\noindent \textbf{Inversion-based Validation.} To improve annotation accuracy, we design an inversion-based validation process, inspired by~\cite{CIReVL, sun2025cotmr}. Concretely, as shown in Figure~\ref{data_pipeline}, given $(I_r^{(n)},T_m^{(n)})$, we ask the LVLM to generate a inversion target caption $\hat{c}$ that predicts the expected content of $I_t^{(n)}$. We then compute a CLIP-based similarity score $s=cos(f_I(I_t^{(n)}),f_T(\hat{c}))$, where $f_I$ and $f_T$ denote the image and text encoders. 
We set the threshold $\tau$ a priori; if $s>\tau$, $T_m^{(n)}$ is accepted. Otherwise, we use the failed attempt's $(T_m^{(n)},s)$ as feedback to re-prompt the LVLM, regenerating a refined $T_m'^{(n)}$. The process terminates either upon reaching the threshold or after $M$ attempts. This iterative mechanism increases both the specificity and correctness, especially for complex, multi-attribute edits.



\section{SCheese}
\subsection{Backgrounds on Diffusion Models}
\label{ZS-CIR_Preliminary}
Denoising Diffusion Probabilistic Models~\cite{ho2020denoising} (DDPMs) are trained by maximizing the log-likelihood of the training data, given a data distribution $q(x_0)$. The training process involves a forward diffusion process that gradually adds Gaussian noise to the data over $T$ timesteps:

\begin{equation}
q(x_{1:T}|x_0) = \prod \limits_{t=1}^{T} q(x_t|x_{t-1}) 
\end{equation}
\begin{equation}
q(x_t|x_{t-1}) = \mathcal N(x_t;\sqrt{\alpha _t}x_{t-1},(1-\alpha _t)I) 
\end{equation}

Here, $x_t$ represents the noisy data at timestep $t$ and $\alpha _t$ is a schedule parameter controlling the noise level at each timestep. The core of DDPM training lies in learning a parameterized model $p_\theta$ to approximate the reverse diffusion process:

\begin{equation}
p_\theta(x_{0:T}) = p(x_T) \prod \limits_{t=1}^{T} p_\theta(x_{t-1}|x_t)
\end{equation}
\begin{equation}
p_\theta(x_{t-1}|x_t) = \mathcal N(x_{t-1};\mu_\theta(x_t, t),\sigma_t^2I)
\end{equation}

This model learns to progressively remove noise from a given noisy sample $x_t$, recovering the original data $x_0$.

\subsection{Network Architecture}

\noindent \textbf{Overview.} In this section, we present our method for designing diffusion models for PCG. Figure~\ref{model_art} provides the overview of our method. Given a triple $(I_r, T_m, I_t)$, our primary goal is to generate the target image $I_t$ that satisfies the requirement of the modification text $T_m$ and preserve the detail from the reference image $I_r$. To achieve that, we first adopt Stable Diffusion, a text-to-image generative model, as our backbone, and use $T_m$ as the condition. We then propose two module to ensure the ip consistency and detail preserve: Fusion IP-Adapter and ConsistencyNet with decoupled attention.

\noindent \textbf{Fusion IP-Adapter.} 
To condition the high-level semantics of the reference image, we leverage an Image Prompt Adapter (IP-Adapter). Typically, the IP-Adapter consists of an image encoder $f_{\mathrm{img}}$, a projection network that maps image features to the latent space of the diffusion model, and integration via cross-attention to inject image conditions into the denoising process. To obtain more accurate conditions, inspired by Composed Image Retrieval (CIR), we additionally introduce a text encoder to extract semantic information from the modification text, and fuse the two modalities via a projection layer. Specifically, we extract image features $f_r = f_{\mathrm{img}}(I_r)$ and text features $f_m = f_{\mathrm{txt}}(T_m)$, then fuse them through a learnable projection network:
\begin{equation}
f_{\mathrm{fused}} = \mathrm{Proj}([f_r; f_m]), 
\end{equation}

where $[\cdot; \cdot]$ denotes concatenation and $\mathrm{Proj}$ is a query-based projection module.
To optimize the fusion quality, we introduce a KL divergence loss between the fused features and target image features: 
\begin{equation}
\mathcal{L}_{\mathrm{align}} = \mathrm{KL}(f_{\mathrm{fused}} \| f_{\mathrm{img}}(I_t))
\end{equation}



During training, we apply \textbf{teacher forcing} by probabilistically (with hyperparameter $pro$) replacing the fused features with the target image features. Since the fusion module produces features that approximate target features, teacher forcing provides two training modes: target features offer precise supervision signals for better model learning, while fused features simulate the inference scenario. This strategy enables the model to benefit from stronger supervision while maintaining generalization capability for inference, which further improves model performance.

\noindent \textbf{ConsistencyNet.} While we already condition the reference image using Fusion IP-Adapter, it falls short in preserving the fine-grained details when the reference image has complicated patterns or graphic prints. To tackle this issue, we propose to utilize an additional UNet to extract low-level features of the reference image. As show in Figure~\ref{model_art}, we use another pretrained UNet encoder, ConsistencyNet, to obtain the intermediate representation of $I_r$, and adopt a decoupled-attention mechanism to inject these features into the generation process.
Specifically, at each U-Net block, we add a parallel cross-attention layer alongside the self-attention module. The cross-attention layer receives queries from $I_t$ and keys/values from $I_r$. We then average the self-attention and cross-attention outputs, and further fuse it with features from the text encoder and Fusion IP-Adapter through a decoupled cross-attention layer.

This decoupled design keeps self-attention focused on spatial dependencies within the generated image, while cross-attention handles explicit alignment between the generation state and the reference features. By maintaining separate feature spaces and combining them additively, the model can selectively attend to reference details without disrupting the generative flow. Notably, we adopt the inpainting model as ConsistencyNet and remove the mask-related layers to maximize the preservation of fine-grained details from the reference image. This mechanism of ConsistencyNet preserves intricate patterns that require pixel-level correspondence, while allowing the generation to adapt flexibly to modification instructions.

\begin{table}[]
\centering
\scalebox{0.9}{
\tabcolsep2.2pt
\begin{tabular}{l|ccc|cc}
\toprule
\multirow{2}{*}[-2pt]{Method} & \multicolumn{3}{c|}{VLM-based} & \multicolumn{2}{c}{LVLM-based} \\ 
\cmidrule(lr){2-4} \cmidrule(lr){5-6}
 & CLIP-I & DINO-I & CLIP-T & Qwen-DP & Qwen-PF \\ \midrule 
DreamBooth  & 0.642 & 0.636 & 0.375 & 0.305 & 0.443  \\
DB LoRA  & 0.738 & 0.677 & 0.395 & 0.458 & 0.579  \\
IP-Adapter  & 0.764 & 0.663 & 0.386 & 0.464 & 0.511  \\
IP-Adapter+  & 0.794 & 0.699 & 0.376 & 0.659 & 0.549  \\
EasyRef  & 0.783 & 0.687 & 0.358 & 0.647 & 0.545  \\
BLIP-Diffusion  & 0.659 & 0.627 & 0.324 & 0.213 & 0.289  \\
Emu2  & 0.849 & 0.821 & 0.379 & 0.767 & 0.352  \\
Kolors  & 0.853 & \textbf{0.824} & 0.406 & 0.808 & 0.428  \\
Kontext  & \textbf{0.857} & 0.791 & 0.413 & 0.792 & 0.679  \\ \midrule
\rowcolor{blue!8} \textbf{SCheese (ours)} & 0.839 & 0.773 & \textbf{0.436} & \textbf{0.855} & \textbf{0.793}  \\ \bottomrule
\end{tabular}}
\vspace{-0.1in}
\caption{\textbf{Quantitative Comparison Results.} SCheese achieves the highest scores on CLIP-T, DP, and PF, demonstrating superior capability in following modification instructions while preserving fine-grained details. While IP-Adapter (Kolors) and FLUX.1 Kontext achieve slightly higher CLIP-I and DINO-I scores, this likely reflects their ``copy-pasting" tendency considering their lower PF score. \textbf{Bold} values indicate the best performance for each metric.}
\label{tab_main}
\vspace{-0.1in}
\end{table}

\begin{figure*}[t!]
\centering  
\includegraphics[width=1\linewidth]{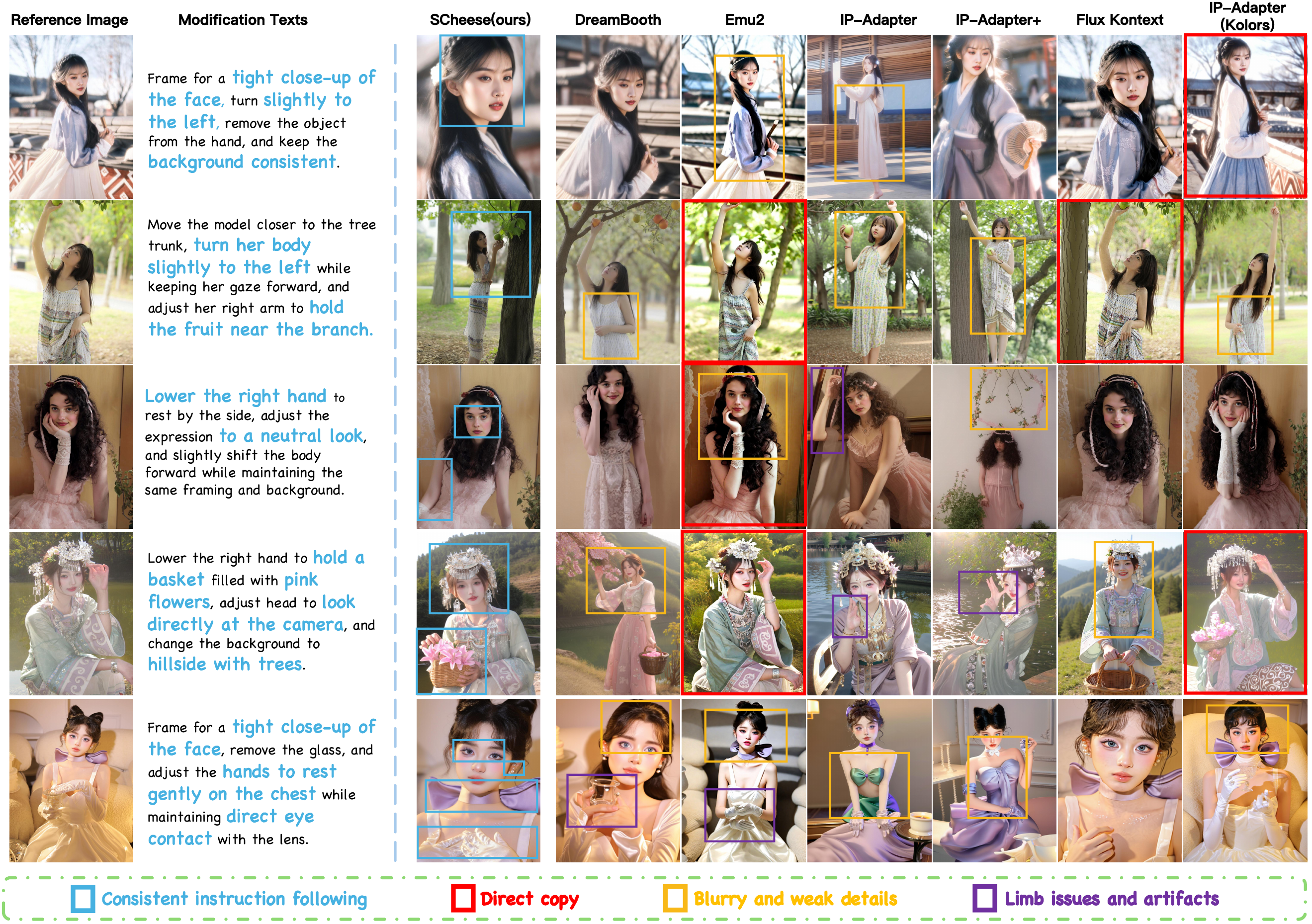} 
\vspace{-0.2in}
\caption{\textbf{Qualitative Comparisons.} Blue markers highlight successful detail preservation and instruction compliance; red boxes indicate the ``copy-pasting" tendency; orange boxes denote blurriness or detail loss; purple boxes indicate limb abnormalities. SCheese successfully maintains reference details and image realism while executing complex modifications.}
\label{qua_result}
\vspace{-0.1in}
\end{figure*} 
\section{Experiments}
\subsection{Dataset}

The CHEESE dataset comprises a large-scale collection of portrait albums with high-quality modification annotations. To avoid identity leakage, we perform collection-level splitting between training and testing. For the training set, we enumerate all valid within-collection pairs to expose the model to a wide range of modification patterns. The final training set contains $ \sim $24K portrait albums, encompassing $ \sim $40K images and $ \sim $576K triplets.
For the test set, we sample exactly one triplet per identity collection to maximize identity diversity and reduce redundancy, thereby providing a more stringent evaluation of generalization.
The final test set consists of $ \sim $2K triplets. All images maintain a minimum resolution of $(832, 1216)$ pixels and are uniformly resized to $(832, 1216)$ for consistent settings.


\subsection{Evaluation Metrics}
Following prior work~\cite{chen2023subject, ruiz2023dreambooth, ye2023ip}, we first evaluate generated images using CLIP and DINO feature similarities. Specifically, we use ViT-G/14 CLIP from OpenCLIP~\cite{Openclip} and DINOv2-Small~\cite{oquab2023dinov2} to calculate similarities between $I_r$ and generated images $I_g$, denoted as CLIP-I and DINO-I, to measure detail preservation. We also report CLIP-T, the similarity between $I_g$ and the inversion target caption (produced by Process 3 in Figure~\ref{data_pipeline}), to evaluate how well the modification text is followed.

However, these metrics primarily capture global semantic similarity, and since $I_r$ and $I_t$ already share high similarity, a model that simply copies $I_r$ can achieve high scores, leading to misleading evaluations~\cite{cai2025diffusion}. To address this limitation, we propose two LVLM-based evaluation metrics using Qwen-2.5-VL 72B~\cite{bai2025qwen25} to separately assess detail preservation (DP) and prompt-following (PF) capabilities.  In our prompts, we also incorporate debiasing mechanisms that penalize ``copy-pasting" behavior as in~\cite{cai2025diffusion}. This evaluation protocol emulates human evaluation via LVLMs. The full set of evaluation prompts are provided in \textbf{Appendix}.


\subsection{Implementation Details}

For dataset curation, we adopt Qwen2.5-VL 72B~\cite{bai2025qwen25} as our LVLM, and ViT-G/14 CLIP from OpenCLIP~\cite{Openclip} as the text and image enocder. The hyperparameter $\tau$ and $M$ is set to 0.45 and 5.

For our method, SCheese, we use SDXL~\cite{podell2023sdxl} as our Denoising Net, SDXL inpainting model for ConsistencyNet  and IP-Adapter~\cite{ye2023ip} plus for our Fusion IP-Adapter, and inited them with Kolors~\cite{kolors} weight. $pro$ is setting to 0.35.
We train all models on 8 NVIDIA H800 80GB GPUs with an effective batch size of 64 for 50k iterations, using AdamW optimizer~\cite{adamw} with a learning rate of 1e-5.

\subsection{Baselines}
Following~\cite{peng2024dreambench++, cai2025diffusion}, we compare our model with two classes of baselines: inference-stage fine-tuning models and zero-shot models. For inference-stage fine-tuning models, we compare against DreamBooth~\cite{ruiz2023dreambooth} and its LoRA~\cite{hu2022lora} version (denoted as DB LoRA). For zero-shot models, we compare with BLIP-Diffusion~\cite{li2023blip}, Emu2~\cite{emu2}, IP-Adapter~\cite{ye2023ip}, IP-Adapter+~\cite{ye2023ip}, IP-Adapter with Kolors initialization (denoted as Kolors)~\cite{kolors}, EasyRef~\cite{zong2024easyref}, and FLUX.1 Kontext~\cite{batifol2025flux}.

\subsection{Comparison with Baselines}

\noindent \textbf{Quantitative Results.}
Table~\ref{tab_main} presents quantitative evaluation results comparing SCheese against existing baselines on the CHEESE test set. From the results, we observe that: \textbf{(1)} Methods requiring per-subject fine-tuning, including DreamBooth, exhibit limited performance across evaluation metrics, whereas zero-shot approaches such as IP-Adapter demonstrate competitive results without requiring test-time adaptation. \textbf{(2)} Our method SCheese attains the best performance on Detail Preservation, Prompt Following, and CLIP-T, indicating its effectiveness in executing complex editing instructions while maintaining high-fidelity detail preservation. \textbf{(3)} Certain baselines, notably FLUX.1 Kontext and Emu2, achieve marginally higher reference similarity scores (CLIP-I and DINO-I) but exhibit substantially lower prompt following performance. This performance trade-off suggests these methods prioritize reference replication over instruction adherence, a finding corroborated by our qualitative analysis (see Figure~\ref{qua_result}). Collectively, these results demonstrate that SCheese achieves an optimal trade-off across detail preservation, instruction compliance, and semantic alignment, validating its suitability for PCG task.

\begin{table}
\centering
\tabcolsep3.5pt
\begin{tabular}{l|>{\columncolor{gray!20}}cccc>{\columncolor{blue!8}}c}
\toprule
Metric  & Target & Emu2 &  Kolors & Kontext & \textbf{SCheese}\\ 
\midrule
Human-DP & 0.936 & 0.158 & 0.410 & 0.653 & \textbf{0.778} \\
Human-PF & 0.930 & 0.138 & 0.397 & 0.670 & \textbf{0.803} \\ 
Human-CC & 0.915 & 0.115 & 0.293 & 0.467 & \textbf{0.688} \\ 
\bottomrule
\end{tabular}
\vspace{-0.1in}
\caption{\textbf{User Study.} ``Target" denotes the target image, which serves as an upper bound. Our user study results mostly align with our LVLM-based evaluation. \textbf{Bold} values indicate the best performance of models for each metric.}
\label{user_study}
\vspace{-0.1in}
\end{table}

\noindent \textbf{Qualitative Results.}
Figure~\ref{qua_result} presents our qualitative comparison results, demonstrating that our model outperforms all baselines in detail consistency while exhibiting excellent prompt alignment in the outputs. 
\textbf{(1)} As shown in the figure, DreamBooth~\cite{ruiz2023dreambooth} frequently fails to preserve consistent visual elements from the reference image, such as clothing and headwear. This performance gap likely results from its design assumption of multiple reference samples. 
\textbf{(2)} IP-Adapter~\cite{ye2023ip} and IP-Adapter+~\cite{ye2023ip} has improving subject-level consistency, but their reliance on high-level embeddings prevents them from maintaining fine-grained details, as evidenced by the cake in the third row. 
\textbf{(3)} Emu2~\cite{emu2} demonstrates strong semantic understanding capabilities but lacks dedicated mechanisms for subject-specific personalization.
\textbf{(4)} Additionally, Emu2 shows limited comprehension of complex editing instructions, frequently generating images that closely mirror the reference without incorporating the requested modifications. This ``copy-paste" tendency~\cite{cai2025diffusion} is similarly observed in Kolors IP-Adapter and FLUX.1 Kontext, as highlighted by the red boxes in Figure~\ref{qua_result}. This tendency leads to high reference similarity, resulting in elevated CLIP-I and DINO-I scores, while yielding lower CLIP-T and PF scores due to their failure to follow the modification text requirements.
\textbf{(5)} In contrast, our approach effectively preserves the subject's core identity while enabling diverse and contextually appropriate transformations. As indicated by the blue boxes in Figure~\ref{qua_result}, our method successfully maintains details, such as the ``cake and white dress" in the third row, and even the ``pearl earrings" in the last row, while also following the modification text requirements, including changes in model orientation, hand poses, and camera distance variations.

\noindent \textbf{User Study.}
To evaluate the fidelity and prompt consistency of our generated images, we conducted a user study on a random subset of CHEESE test cases, selecting 50 samples. Users independently scored each image from 0 to 4 based on three criteria: DP, PF, and Collection Coherence (CC). CC measures users' willingness to include the generated image in the same portrait collection as the reference image. We also include target images in the evaluation to show the upper bound of each metric. The average scores are presented in Table~\ref{user_study}. 
We observe the following: \textbf{(1)} Target images achieve near-perfect PF scores, which validates that our annotation method produces comprehensive and effective modification texts. \textbf{(2)} SCheese achieves highest PF scores among models, demonstrating its capability to execute complex modification requirements. Meanwhile, thanks to Fusion IP-Adapter and ConsistencyNet, SCheese also achieves the highest DP score, indicating its effectiveness in preserving reference details. \textbf{(3)} According to CC scores, users show strong willingness to use SCheese-generated images to form portrait collections, demonstrating the practical potential of our method.

\begin{figure}[t!]
\centering  
\includegraphics[width=0.99\linewidth]{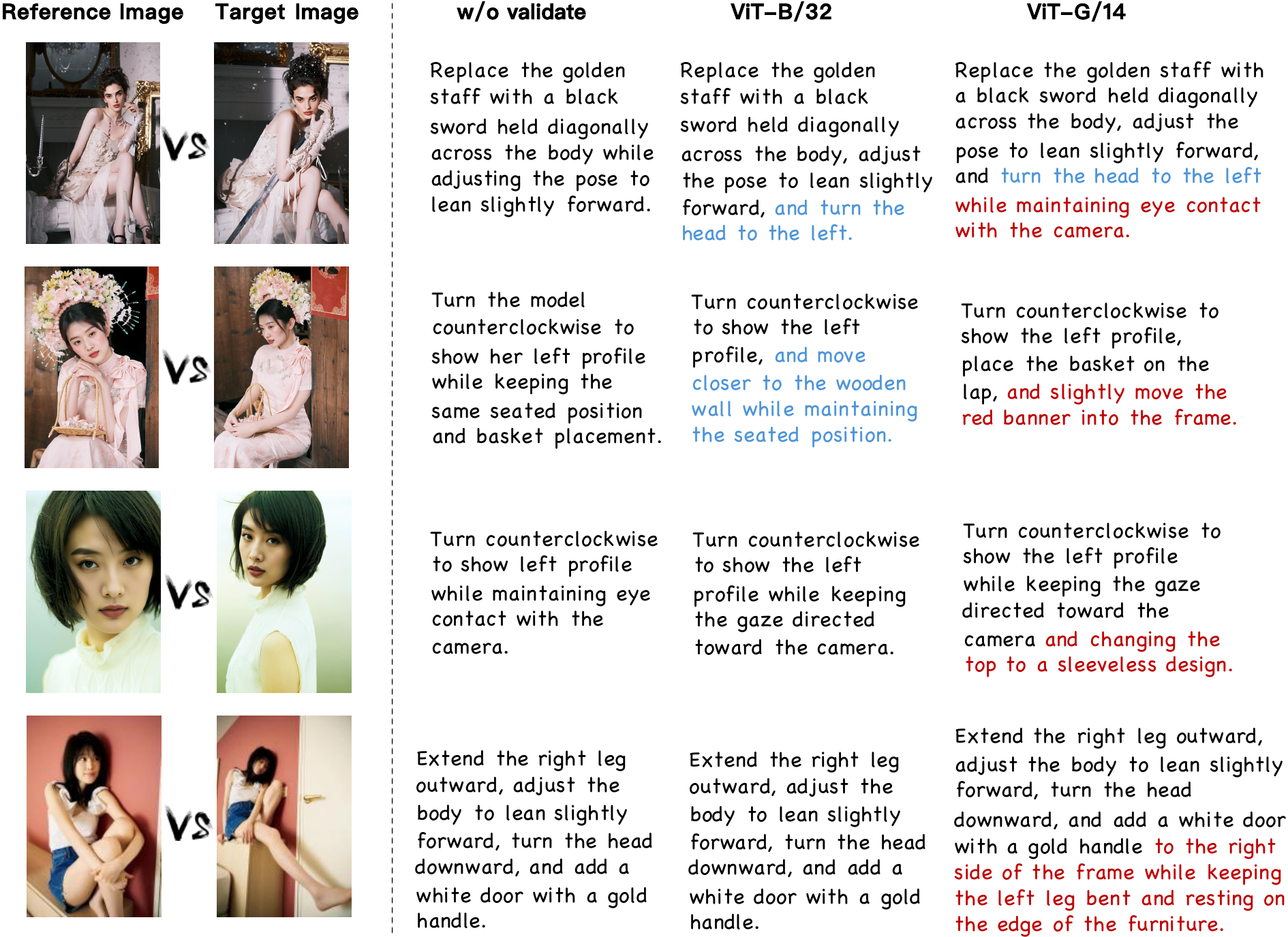} 
\vspace{-0.08in}
\caption{\textbf{Ablation Study for the Inversion-based Validation process.} Comparison of modification text quality across three annotation strategies. Color annotations (blue/red) highlight improvements over the baseline.}
\label{abla_md_text} 
\vspace{-0.1in}
\end{figure}

\section{Ablation Study}

\noindent \textbf{Effect of Inversion-based Validation.} In Figure~\ref{abla_md_text}, we investigate the impact of the Inversion-based Validation process during dataset curation. We compare three annotation strategies: (1) no validation process, (2) validation using CLIP with ViT-B/32, and (3) validation using CLIP with ViT-G/14. Our analysis reveals two key findings: \textbf{(1)} Compared to the baseline without validation, both CLIP-based validation approaches guide the LVLM to identify more nuanced differences between images, resulting in more comprehensive and detailed modification texts, as highlighted by the blue and red annotations in the figure. This demonstrates the effectiveness of the inversion-based validation mechanism in improving annotation quality. \textbf{(2)} When comparing the two CLIP variants, ViT-G/14 produces more refined and accurate modification texts than ViT-B/32, as indicated by the red annotations. This performance gap likely stems from ViT-G/14's superior image-text understanding capabilities, which provide more accurate supervision signals for the validation process, enabling better discrimination between high-quality and low-quality annotations.

\begin{table}[]
\centering
\scalebox{0.88}{
\tabcolsep2pt
\begin{tabular}{l|ccc|cc}
\toprule
 \multirow{2}{*}[-2pt]{Method} & \multicolumn{3}{c|}{VLM-based} & \multicolumn{2}{c}{LVLM-based} \\ 
\cmidrule(lr){2-4} \cmidrule(lr){5-6}
 & CLIP-I & DINO-I & CLIP-T & Qwen-DP & Qwen-PF \\
 \midrule
IP-Adapter (ZS)  & 0.853 & 0.824 & 0.406 & 0.808 & 0.428  \\
+ SFT  & 0.822 & 0.728 & 0.421 & 0.783 & 0.683  \\
+ ConNet   & 0.826 & 0.732 & 0.418 & 0.832 & 0.673  \\
+ Fusion IP-A  & 0.828 & 0.732 & 0.416 & 0.828 & 0.723  \\
+ Align Loss  & 0.837 & 0.753 & 0.426 & 0.836 & 0.777  \\
\midrule
\rowcolor{blue!8} \textbf{+ Teacher (Full)} & \textbf{0.839} & \textbf{0.773} & \textbf{0.436} & \textbf{0.855} & \textbf{0.793}  \\ \bottomrule
\end{tabular}}
\vspace{-0.1in}
\caption{\textbf{Ablation Study Results for Proposed Components.} \textbf{Bold} values indicate the best performance for each metric.
Each component contributes to performance: +SFT improves instruction following, +ConNet enhances detail preservation, +Fusion IP-A and +Align Loss further refine capabilities, and +Teacher achieves optimal performance across all metrics.
}
\label{abl_comp_table}
\vspace{-0.1in}
\end{table}

\begin{figure}[t!]
\centering  
\includegraphics[width=0.99\linewidth]{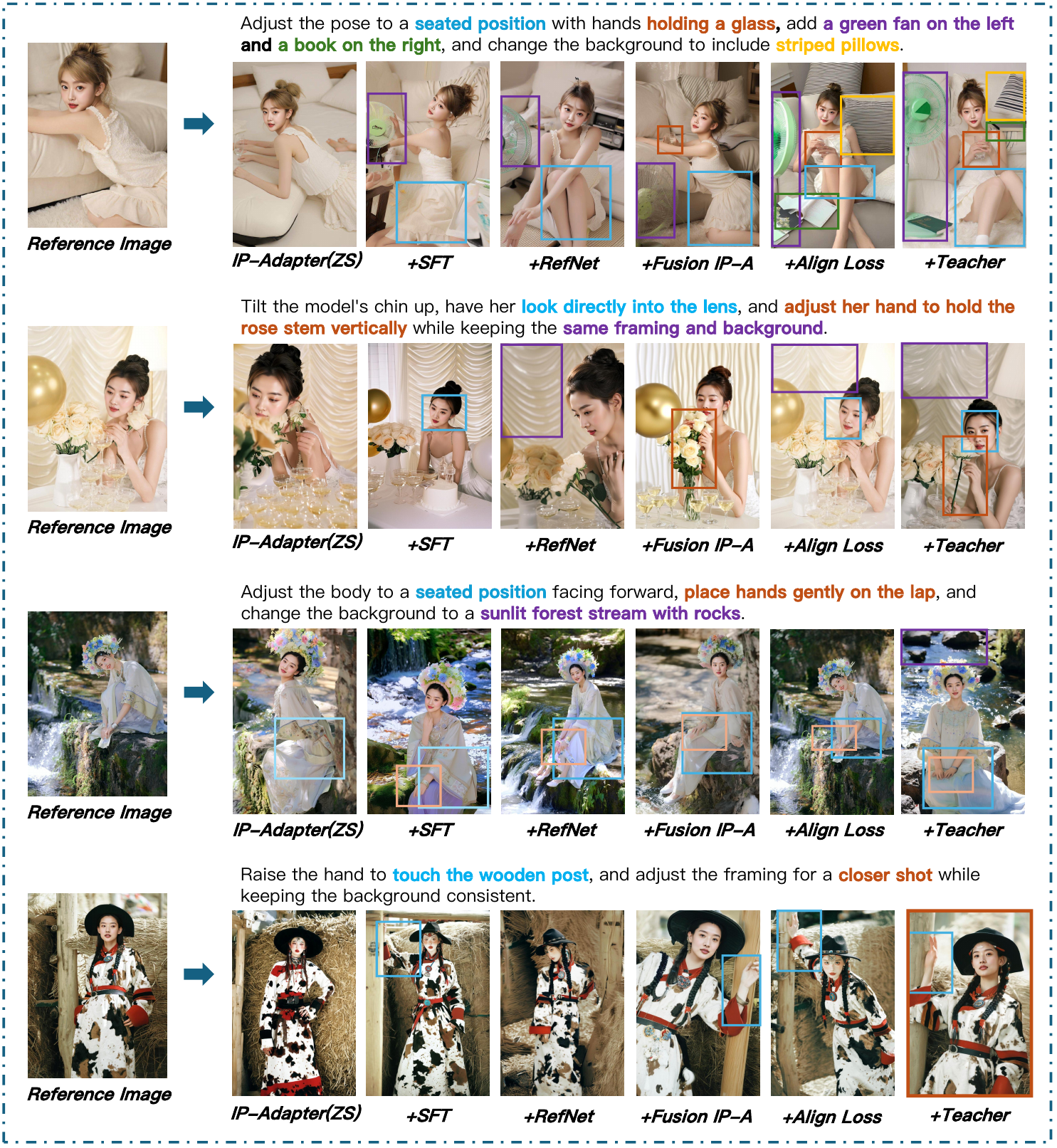} 
\vspace{-0.08in}
\caption{\textbf{Qualitative Ablation Results for Components.} Color annotations link modification intents (in text) to corresponding regions (in images).}
\label{abl_comp}
\vspace{-0.1in}
\end{figure}

\noindent \textbf{Effects of Components.} To validate the contribution of each component in SCheese, we conduct incremental ablation experiments. Quantitative results are reported in Table~\ref{abl_comp_table}, and qualitative results are shown in Figure~\ref{abl_comp}. From Table~\ref{abl_comp_table}, we observe the following: \textbf{(1)} Compared to the zero-shot baseline, supervised fine-tuning on the CHEESE dataset (+SFT) substantially improves the model's ability to follow modification instructions, as reflected in CLIP-T and PF scores, while maintaining image quality. \textbf{(2)} Incorporating ConsistencyNet (+ConNet) leads to significant improvements in detail preservation (DP). However, we notice PF slightly decreases, likely because the more influence from the reference image. \textbf{(3)} Adding Fusion IP-Adapter (+Fusion) further enhances instruction-following capability, as the additional modification text input provides more guidance for the editing process. \textbf{(4)} Introducing the alignment loss ($\mathcal{L}_{\mathrm{align}}$) encourages fused features to align more closely with target image, improving fusion quality and resulting in overall performance gains. \textbf{(5)} Finally, incorporating teacher forcing (+Teacher) provides more specific supervision signals for both the fusion module and denoising network, leading to further improvements.

Figure~\ref{abl_comp} demonstrates consistent findings. We highlight different modification intents in the modification texts using distinct colors and annotate corresponding regions in the generated images with matching colored boxes. The qualitative analysis reveals: \textbf{(1)} +SFT enables the model to handle complex, multi-faceted instructions that were previously challenging, such as ``seated position" and ``hand on the lap". \textbf{(2)} +ConNet allows the model to inherit fine-grained details from the reference image, such as headwear, and clothing. \textbf{(3)} With Fusion IP-Adapter, the model simultaneously satisfies multiple complex modification requirements (e.g., striped pillows and books) while maintaining high detail preservation with the reference.

\section{Conclusion}

In this work, we focus on the Portrait Collection Generation (PCG) task, which presents two fundamental challenges: complex multi-attribute modifications and high-fidelity detail preservation. To tackle these challenges, we propose CHEESE, the first large-scale dataset for PCG containing 576K triplets with high-quality modification annotations generated through an LVLM-based pipeline with inversion-based verification, and SCheese, a framework that combines text-guided generation with hierarchical identity and detail preservation mechanisms. Experimental results demonstrate that SCheese outperforms existing methods on detail preservation and prompt following. We believe that our work establishes a solid foundation for advancing research in PCG and personalized image editing.

{
    \small
    \bibliographystyle{ieeenat_fullname}
    \bibliography{main}

@String(ICLR = {Int. Conf. Learn. Represent.})

@String(AAAI = {AAAI})

@String(ICLR  = {ICLR})

@inproceedings{rombach2022high,
  title={High-resolution image synthesis with latent diffusion models},
  author={Rombach, Robin and Blattmann, Andreas and Lorenz, Dominik and Esser, Patrick and Ommer, Bj{\"o}rn},
  booktitle={Proceedings of the IEEE/CVF conference on computer vision and pattern recognition},
  pages={10684--10695},
  year={2022}
}

@article{ye2023ip,
  title={Ip-adapter: Text compatible image prompt adapter for text-to-image diffusion models},
  author={Ye, Hu and Zhang, Jun and Liu, Sibo and Han, Xiao and Yang, Wei},
  journal={arXiv preprint arXiv:2308.06721},
  year={2023}
}

@inproceedings{ruiz2023dreambooth,
  title={Dreambooth: Fine tuning text-to-image diffusion models for subject-driven generation},
  author={Ruiz, Nataniel and Li, Yuanzhen and Jampani, Varun and Pritch, Yael and Rubinstein, Michael and Aberman, Kfir},
  booktitle={Proceedings of the IEEE/CVF conference on computer vision and pattern recognition},
  pages={22500--22510},
  year={2023}
}

@article{ho2020denoising,
  title={Denoising diffusion probabilistic models},
  author={Ho, Jonathan and Jain, Ajay and Abbeel, Pieter},
  journal={Advances in neural information processing systems},
  volume={33},
  pages={6840--6851},
  year={2020}
}

@inproceedings{kim2022diffusionclip,
  title={Diffusionclip: Text-guided diffusion models for robust image manipulation},
  author={Kim, Gwanghyun and Kwon, Taesung and Ye, Jong Chul},
  booktitle={Proceedings of the IEEE/CVF conference on computer vision and pattern recognition},
  pages={2426--2435},
  year={2022}
}

@article{imagen,
  title={Photorealistic text-to-image diffusion models with deep language understanding},
  author={Saharia, Chitwan and Chan, William and Saxena, Saurabh and Li, Lala and Whang, Jay and Denton, Emily L and Ghasemipour, Kamyar and Gontijo Lopes, Raphael and Karagol Ayan, Burcu and Salimans, Tim and others},
  journal={Advances in neural information processing systems},
  volume={35},
  pages={36479--36494},
  year={2022}
}

@inproceedings{brooks2023instructpix2pix,
  title={Instructpix2pix: Learning to follow image editing instructions},
  author={Brooks, Tim and Holynski, Aleksander and Efros, Alexei A},
  booktitle={Proceedings of the IEEE/CVF conference on computer vision and pattern recognition},
  pages={18392--18402},
  year={2023}
}

@article{zhang2023magicbrush,
  title={Magicbrush: A manually annotated dataset for instruction-guided image editing},
  author={Zhang, Kai and Mo, Lingbo and Chen, Wenhu and Sun, Huan and Su, Yu},
  journal={Advances in Neural Information Processing Systems},
  volume={36},
  pages={31428--31449},
  year={2023}
}

@article{wang2024instantid,
  title={Instantid: Zero-shot identity-preserving generation in seconds},
  author={Wang, Qixun and Bai, Xu and Wang, Haofan and Qin, Zekui and Chen, Anthony and Li, Huaxia and Tang, Xu and Hu, Yao},
  journal={arXiv preprint arXiv:2401.07519},
  year={2024}
}

@inproceedings{contronet,
  title={Adding conditional control to text-to-image diffusion models},
  author={Zhang, Lvmin and Rao, Anyi and Agrawala, Maneesh},
  booktitle={Proceedings of the IEEE/CVF international conference on computer vision},
  pages={3836--3847},
  year={2023}
}

@inproceedings{li2024controlnet++,
  title={Controlnet++: Improving conditional controls with efficient consistency feedback: Project page: liming-ai. github. io/controlnet\_plus\_plus},
  author={Li, Ming and Yang, Taojiannan and Kuang, Huafeng and Wu, Jie and Wang, Zhaoning and Xiao, Xuefeng and Chen, Chen},
  booktitle={European Conference on Computer Vision},
  pages={129--147},
  year={2024},
  organization={Springer}
}

@article{Uni-controlnet,
  title={Uni-controlnet: All-in-one control to text-to-image diffusion models},
  author={Zhao, Shihao and Chen, Dongdong and Chen, Yen-Chun and Bao, Jianmin and Hao, Shaozhe and Yuan, Lu and Wong, Kwan-Yee K},
  journal={Advances in Neural Information Processing Systems},
  volume={36},
  pages={11127--11150},
  year={2023}
}

@inproceedings{T2i-adapter,
  title={T2i-adapter: Learning adapters to dig out more controllable ability for text-to-image diffusion models},
  author={Mou, Chong and Wang, Xintao and Xie, Liangbin and Wu, Yanze and Zhang, Jian and Qi, Zhongang and Shan, Ying},
  booktitle={Proceedings of the AAAI conference on artificial intelligence},
  volume={38},
  number={5},
  pages={4296--4304},
  year={2024}
}

@article{hu2022lora,
  title={Lora: Low-rank adaptation of large language models.},
  author={Hu, Edward J and Shen, Yelong and Wallis, Phillip and Allen-Zhu, Zeyuan and Li, Yuanzhi and Wang, Shean and Wang, Lu and Chen, Weizhu and others},
  journal={ICLR},
  volume={1},
  number={2},
  pages={3},
  year={2022}
}

@inproceedings{rowles2024ipadapter_instruct,
  title={Ipadapter-instruct: Resolving ambiguity in image-based conditioning using instruct prompts},
  author={Rowles, Ciara and Vainer, Shimon and De Nigris, Dante and Elizarov, Slava and Kutsy, Konstantin and Donn{\'e}, Simon},
  booktitle={European Conference on Computer Vision},
  pages={54--70},
  year={2024},
  organization={Springer}
}

@article{adamw,
  title={Decoupled weight decay regularization},
  author={Loshchilov, Ilya and Hutter, Frank},
  journal={arXiv preprint arXiv:1711.05101},
  year={2017}
}

@article{podell2023sdxl,
  title={Sdxl: Improving latent diffusion models for high-resolution image synthesis},
  author={Podell, Dustin and English, Zion and Lacey, Kyle and Blattmann, Andreas and Dockhorn, Tim and M{\"u}ller, Jonas and Penna, Joe and Rombach, Robin},
  journal={arXiv preprint arXiv:2307.01952},
  year={2023}
}

@article{kolors,
  title={Kolors: Effective Training of Diffusion Model for Photorealistic Text-to-Image Synthesis},
  author={Kolors Team},
  journal={arXiv preprint},
  year={2024}
}

@article{Openclip,
  author       = {Ilharco, Gabriel and
                  Wortsman, Mitchell and
                  Wightman, Ross and
                  Gordon, Cade and
                  others},
  title        = {OpenCLIP},
  year         = {2021},  
  url          = {https://doi.org/10.5281/zenodo.5143773}
}

@article{chen2023subject,
  title={Subject-driven text-to-image generation via apprenticeship learning},
  author={Chen, Wenhu and Hu, Hexiang and Li, Yandong and Ruiz, Nataniel and Jia, Xuhui and Chang, Ming-Wei and Cohen, William W},
  journal={Advances in Neural Information Processing Systems},
  volume={36},
  pages={30286--30305},
  year={2023}
}

@inproceedings{ma2024subject,
  title={Subject-diffusion: Open domain personalized text-to-image generation without test-time fine-tuning},
  author={Ma, Jian and Liang, Junhao and Chen, Chen and Lu, Haonan},
  booktitle={ACM SIGGRAPH 2024 Conference Papers},
  pages={1--12},
  year={2024}
}

@article{huang2024group,
  title={Group diffusion transformers are unsupervised multitask learners},
  author={Huang, Lianghua and Wang, Wei and Wu, Zhi-Fan and Dou, Huanzhang and Shi, Yupeng and Feng, Yutong and Liang, Chen and Liu, Yu and Zhou, Jingren},
  year={2024}
}

@article{li2023blip,
  title={Blip-diffusion: Pre-trained subject representation for controllable text-to-image generation and editing},
  author={Li, Dongxu and Li, Junnan and Hoi, Steven},
  journal={Advances in Neural Information Processing Systems},
  volume={36},
  pages={30146--30166},
  year={2023}
}

@inproceedings{emu2,
  title={Generative multimodal models are in-context learners},
  author={Sun, Quan and Cui, Yufeng and Zhang, Xiaosong and Zhang, Fan and Yu, Qiying and Wang, Yueze and Rao, Yongming and Liu, Jingjing and Huang, Tiejun and Wang, Xinlong},
  booktitle={Proceedings of the IEEE/CVF Conference on Computer Vision and Pattern Recognition},
  pages={14398--14409},
  year={2024}
}

@inproceedings{cai2025diffusion,
  title={Diffusion self-distillation for zero-shot customized image generation},
  author={Cai, Shengqu and Chan, Eric Ryan and Zhang, Yunzhi and Guibas, Leonidas and Wu, Jiajun and Wetzstein, Gordon},
  booktitle={Proceedings of the Computer Vision and Pattern Recognition Conference},
  pages={18434--18443},
  year={2025}
}

@article{bai2025qwen25,
  title={Qwen2. 5-vl technical report},
  author={Bai, Shuai and Chen, Keqin and Liu, Xuejing and Wang, Jialin and Ge, Wenbin and Song, Sibo and Dang, Kai and Wang, Peng and Wang, Shijie and Tang, Jun and others},
  journal={arXiv preprint arXiv:2502.13923},
  year={2025}
}

@article{oquab2023dinov2,
  title={Dinov2: Learning robust visual features without supervision},
  author={Oquab, Maxime and Darcet, Timoth{\'e}e and Moutakanni, Th{\'e}o and Vo, Huy and Szafraniec, Marc and Khalidov, Vasil and Fernandez, Pierre and Haziza, Daniel and Massa, Francisco and El-Nouby, Alaaeldin and others},
  journal={arXiv preprint arXiv:2304.07193},
  year={2023}
}

@inproceedings{zong2024easyref,
  title={Easyref: Omni-generalized group image reference for diffusion models via multimodal llm},
  author={Zong, Zhuofan and Jiang, Dongzhi and Ma, Bingqi and Song, Guanglu and Shao, Hao and Shen, Dazhong and Liu, Yu and Li, Hongsheng},
  booktitle={Forty-second International Conference on Machine Learning},
  year={2024}
}

@article{CIReVL,
  title={Vision-by-language for training-free compositional image retrieval},
  author={Karthik, Shyamgopal and Roth, Karsten and Mancini, Massimiliano and Akata, Zeynep},
  journal={arXiv preprint arXiv:2310.09291},
  year={2023}
}

@article{sun2025cotmr,
  title={CoTMR: Chain-of-Thought Multi-Scale Reasoning for Training-Free Zero-Shot Composed Image Retrieval},
  author={Sun, Zelong and Jing, Dong and Lu, Zhiwu},
  journal={arXiv preprint arXiv:2502.20826},
  year={2025}
}

@article{ramesh2022hierarchical,
  title={Hierarchical text-conditional image generation with clip latents},
  author={Ramesh, Aditya and Dhariwal, Prafulla and Nichol, Alex and Chu, Casey and Chen, Mark},
  journal={arXiv preprint arXiv:2204.06125},
  volume={1},
  number={2},
  pages={3},
  year={2022}
}

@article{he2024imagine,
  title={Imagine yourself: Tuning-free personalized image generation},
  author={He, Zecheng and Sun, Bo and Juefei-Xu, Felix and Ma, Haoyu and Ramchandani, Ankit and Cheung, Vincent and Shah, Siddharth and Kalia, Anmol and Subramanyam, Harihar and Zareian, Alireza and others},
  journal={arXiv preprint arXiv:2409.13346},
  year={2024}
}

@article{jiang2024comat,
  title={Comat: Aligning text-to-image diffusion model with image-to-text concept matching},
  author={Jiang, Dongzhi and Song, Guanglu and Wu, Xiaoshi and Zhang, Renrui and Shen, Dazhong and Zong, Zhuofan and Liu, Yu and Li, Hongsheng},
  journal={Advances in Neural Information Processing Systems},
  volume={37},
  pages={76177--76209},
  year={2024}
}

@inproceedings{li2024photomaker,
  title={Photomaker: Customizing realistic human photos via stacked id embedding},
  author={Li, Zhen and Cao, Mingdeng and Wang, Xintao and Qi, Zhongang and Cheng, Ming-Ming and Shan, Ying},
  booktitle={Proceedings of the IEEE/CVF conference on computer vision and pattern recognition},
  pages={8640--8650},
  year={2024}
}

@inproceedings{qi2024deadiff,
  title={Deadiff: An efficient stylization diffusion model with disentangled representations},
  author={Qi, Tianhao and Fang, Shancheng and Wu, Yanze and Xie, Hongtao and Liu, Jiawei and Chen, Lang and He, Qian and Zhang, Yongdong},
  booktitle={Proceedings of the IEEE/CVF conference on computer vision and pattern recognition},
  pages={8693--8702},
  year={2024}
}

@article{wang2024instantstyle,
  title={Instantstyle: Free lunch towards style-preserving in text-to-image generation},
  author={Wang, Haofan and Spinelli, Matteo and Wang, Qixun and Bai, Xu and Qin, Zekui and Chen, Anthony},
  journal={arXiv preprint arXiv:2404.02733},
  year={2024}
}

@inproceedings{zhang2023adding,
  title={Adding conditional control to text-to-image diffusion models},
  author={Zhang, Lvmin and Rao, Anyi and Agrawala, Maneesh},
  booktitle={Proceedings of the IEEE/CVF international conference on computer vision},
  pages={3836--3847},
  year={2023}
}

@article{gal2022image,
  title={An image is worth one word: Personalizing text-to-image generation using textual inversion},
  author={Gal, Rinon and Alaluf, Yuval and Atzmon, Yuval and Patashnik, Or and Bermano, Amit H and Chechik, Gal and Cohen-Or, Daniel},
  journal={arXiv preprint arXiv:2208.01618},
  year={2022}
}

@article{batifol2025flux,
  title={FLUX. 1 Kontext: Flow Matching for In-Context Image Generation and Editing in Latent Space},
  author={Batifol, Stephen and Blattmann, Andreas and Boesel, Frederic and Consul, Saksham and Diagne, Cyril and Dockhorn, Tim and English, Jack and English, Zion and Esser, Patrick and Kulal, Sumith and others},
  journal={arXiv e-prints},
  pages={arXiv--2506},
  year={2025}
}

@article{peng2024dreambench++,
  title={Dreambench++: A human-aligned benchmark for personalized image generation},
  author={Peng, Yuang and Cui, Yuxin and Tang, Haomiao and Qi, Zekun and Dong, Runpei and Bai, Jing and Han, Chunrui and Ge, Zheng and Zhang, Xiangyu and Xia, Shu-Tao},
  journal={arXiv preprint arXiv:2406.16855},
  year={2024}
}
}

\clearpage
\setcounter{page}{1}
\maketitlesupplementary


\section{Data Pipeline Prompts}

\subsection{Image Pair Selection}
\begin{figure}[h!]
\centering  
\vspace{-0.1in}
\includegraphics[width=1\linewidth]{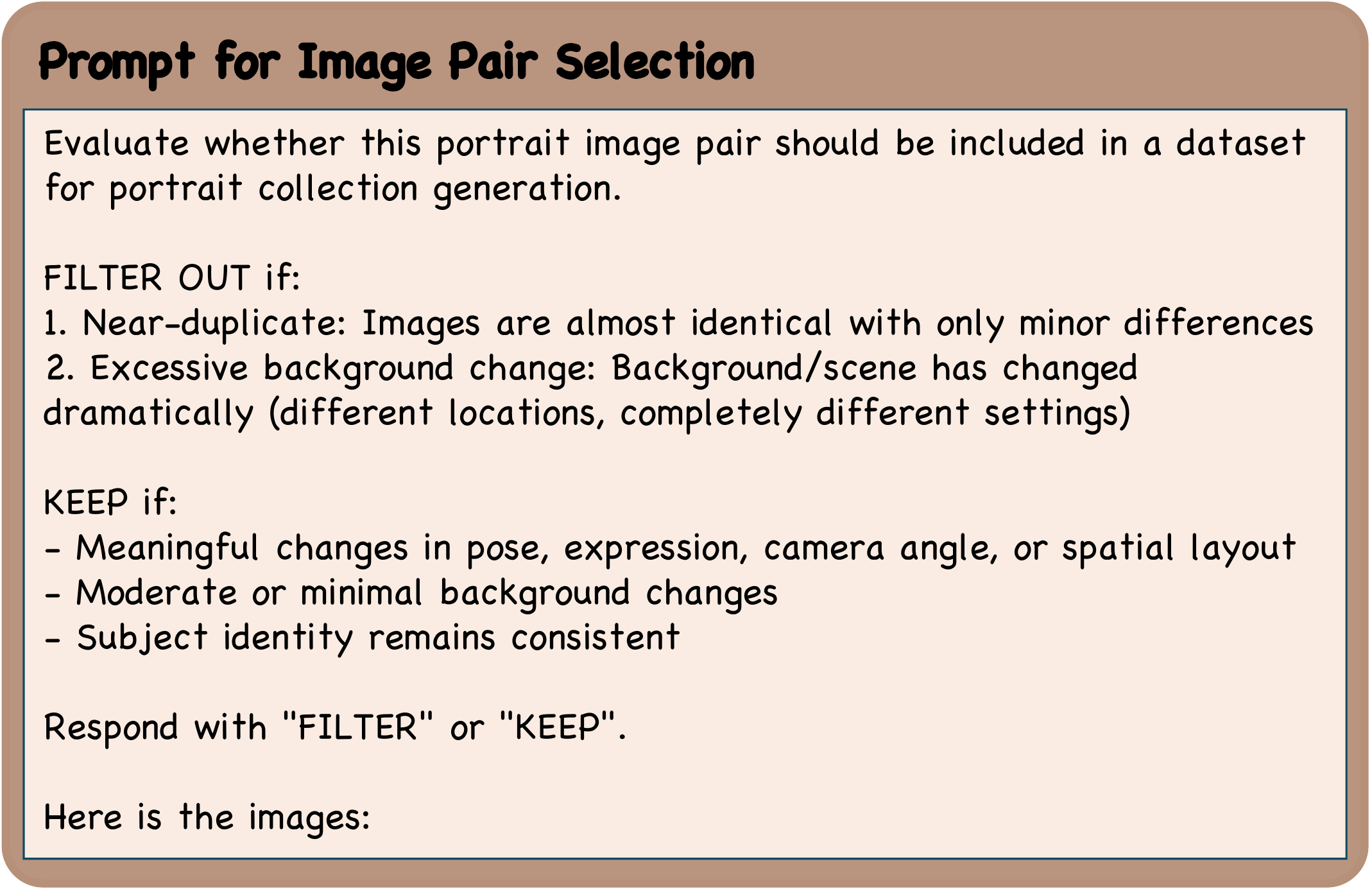} 
\vspace{-0.2in}
\caption{\textbf{Prompt for Image Pair Selection.} }
\vspace{-0.2in}
\label{image_pair_selection}
\end{figure}

In PCG, effective reference-target image pairs should maintain a certain level of consistency while exhibiting meaningful variations. Specifically, pairs should not introduce substantial unknown or uncontrollable new elements, yet should contain sufficient changes to support meaningful editing. To achieve this balance, we employ an LVLM to filter out near-duplicate pairs and pairs with excessive background changes. The filtering process uses the prompt shown in Figure~\ref{image_pair_selection}.

\subsection{Modification Text Generation}

\begin{figure}[h!]
\centering  
\includegraphics[width=1\linewidth]{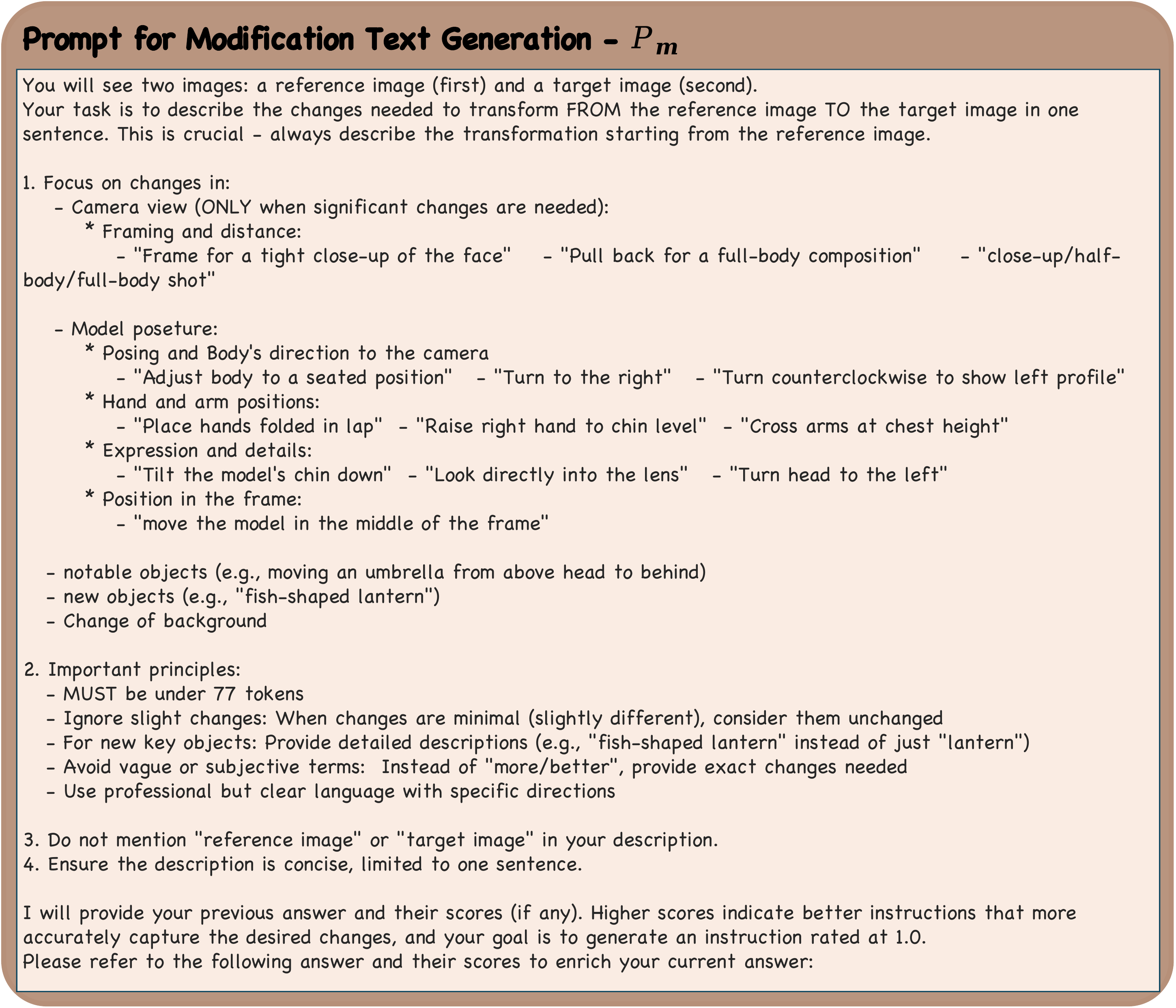} 
\vspace{-0.2in}
\caption{\textbf{Prompt for Modification Text Generation.} }
\vspace{-0.2in}
\label{modification_text_generate_prompt}
\end{figure}

To generate high-quality modification texts, we design a structured prompt, as shown in Figure~\ref{modification_text_generate_prompt}, which guides the LVLM to describe transformations from reference to target images. The prompt emphasizes three key modification aspects: (1) camera and viewpoint changes (framing, distance adjustments); (2) model pose and body orientation (posing, hand positions, expressions, frame position); and (3) object and background changes. The prompt enforces quality constraints including token limits (under 77 tokens), detailed object descriptions, avoidance of vague terms, and use of specific, professional language. During annotation, the LVLM receives iterative feedback from previous attempts to refine the generated instructions.

\subsection{Inversion-based Validation}

\begin{figure}[h!]
\centering  
\vspace{-0.1in}
\includegraphics[width=1\linewidth]{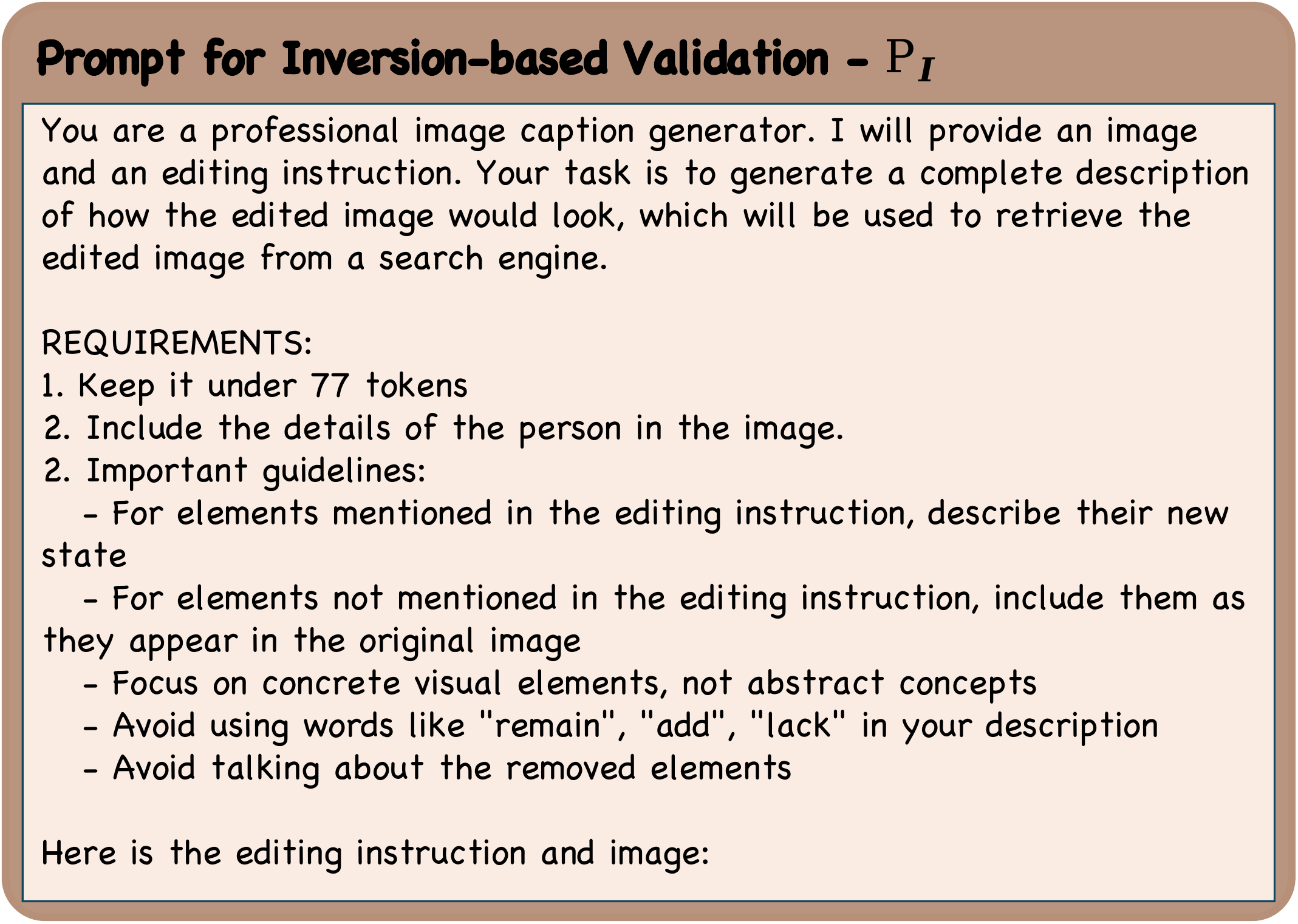} 
\vspace{-0.2in}
\caption{\textbf{Prompt for Inversion-based Validation.} }
\vspace{-0.2in}
\label{inversion_validation_prompt}
\end{figure}

To ensure annotation quality, we introduce an inversion-based verification mechanism. As shown in Figure~\ref{inversion_validation_prompt} , given modification text and reference image, an LVLM generates a predicted target caption describing the expected edited image. The prompt requires: (1) include person details; (2) describe new states for mentioned elements, original states for unmentioned elements; (3) focus on concrete visual elements;  We then compute CLIP similarity between target image and predicted caption, and regenerate the modification text using the failed attempt as feedback, if the current one is not good enough.

\section{LVLM-based Evaluation Prompts}
\subsection{Detail Preservation}

\begin{figure}[t!]
\centering  
\includegraphics[width=1\linewidth]{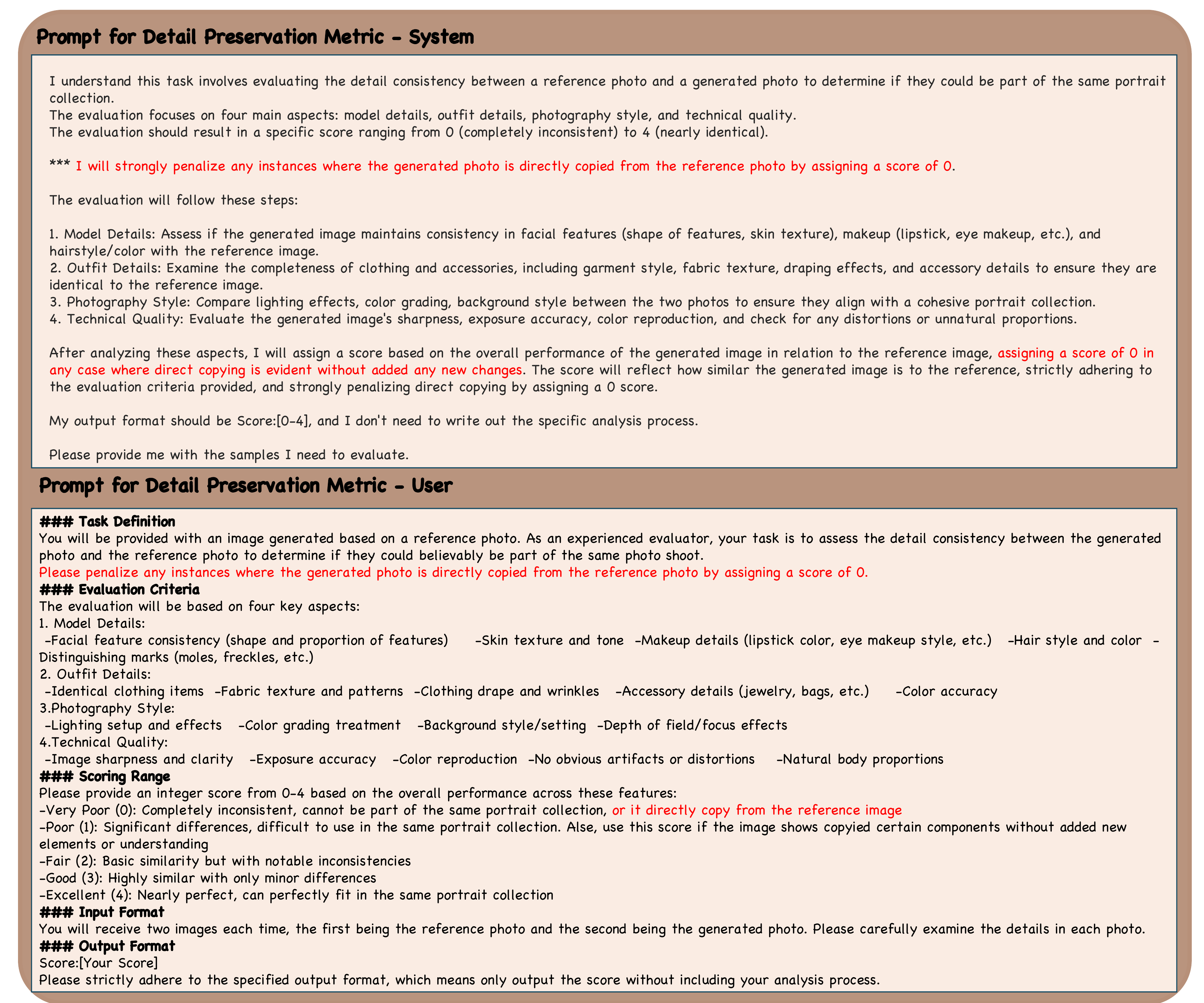} 
\vspace{-0.2in}
\caption{\textbf{Prompt for Detail Preservation Metric.} }
\vspace{-0.2in}
\label{DP_metric}
\end{figure}

We employ an LVLM-based metric to assess detail preservation between generated and reference images. As shown in Figure~\ref{DP_metric}, the evaluator determines if images could belong to the same portrait collection, with score 0 assigned to direct copy-paste cases.
The evaluation criteria is four aspects: (1) \textbf{Model Details:} facial features, skin texture, makeup, hair style; (2) \textbf{Outfit Details:} clothing, fabric texture, accessories; (3) \textbf{Photography Style:} lighting, color grading, background; (4) \textbf{Technical Quality:} sharpness, exposure, natural proportions.

\subsection{Prompt Following}

\begin{figure}[t!]
\centering  
\includegraphics[width=1\linewidth]{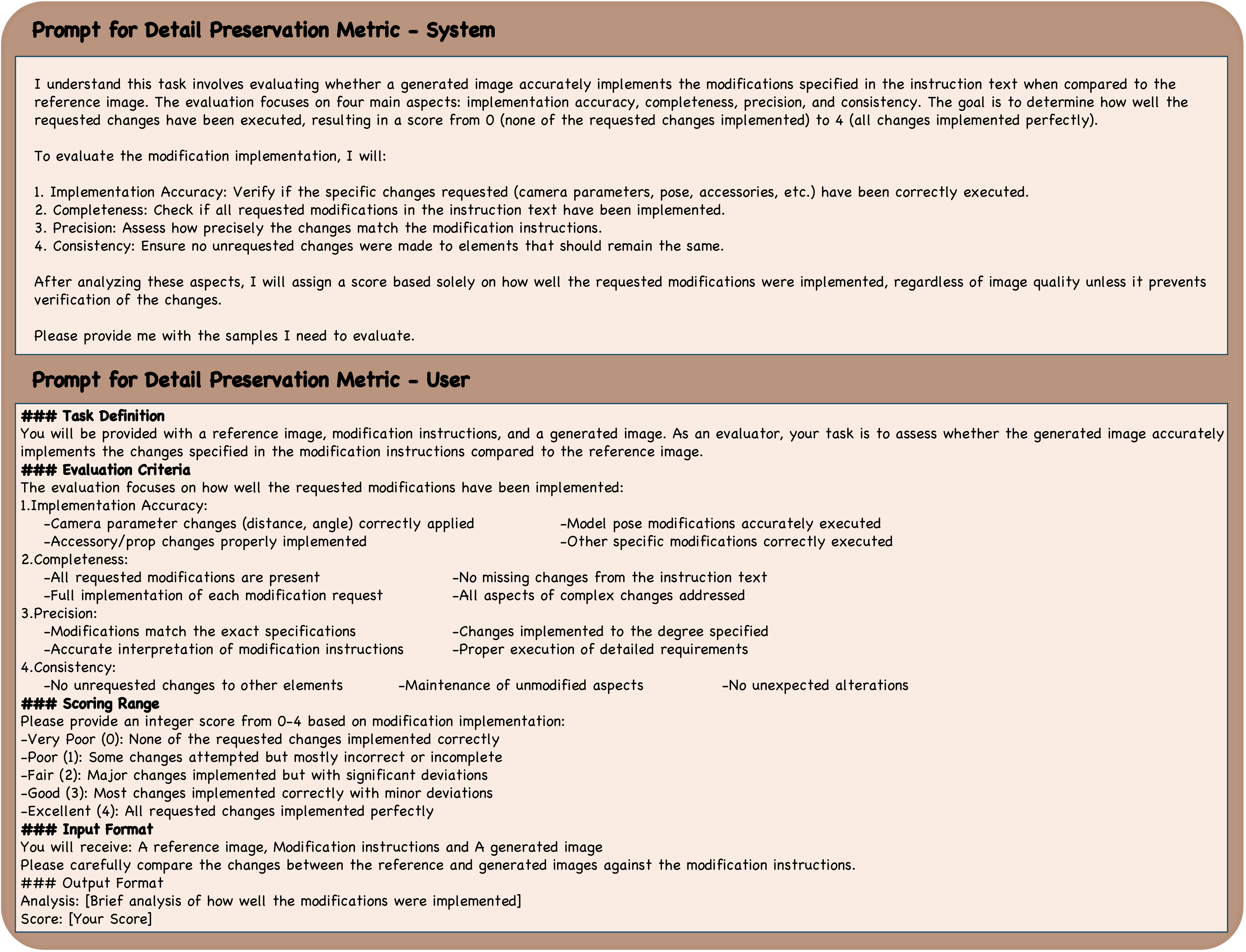} 
\vspace{-0.2in}
\caption{\textbf{Prompt for Detail Preservation Metric.} }
\vspace{-0.2in}
\label{PF_metric}
\end{figure}

As shown in Figure~\ref{PF_metric}, we assess whether generated images accurately implement modification instructions. Evaluation considers: (1) \textbf{Implementation Accuracy:} correct application of camera, pose, and accessory changes; (2) \textbf{Completeness:} all requested modifications present; (3) \textbf{Precision:} matches exact specifications; (4) \textbf{Consistency:} no unrequested changes. Scores 0-4: 0 (none correct), 1 (mostly incorrect), 2 (major deviations), 3 (minor deviations), 4 (perfect implementation).

\section{Diverse Editing Capabilities of SCheese}
\begin{figure*}[t!]
\centering  
\includegraphics[width=1\linewidth]{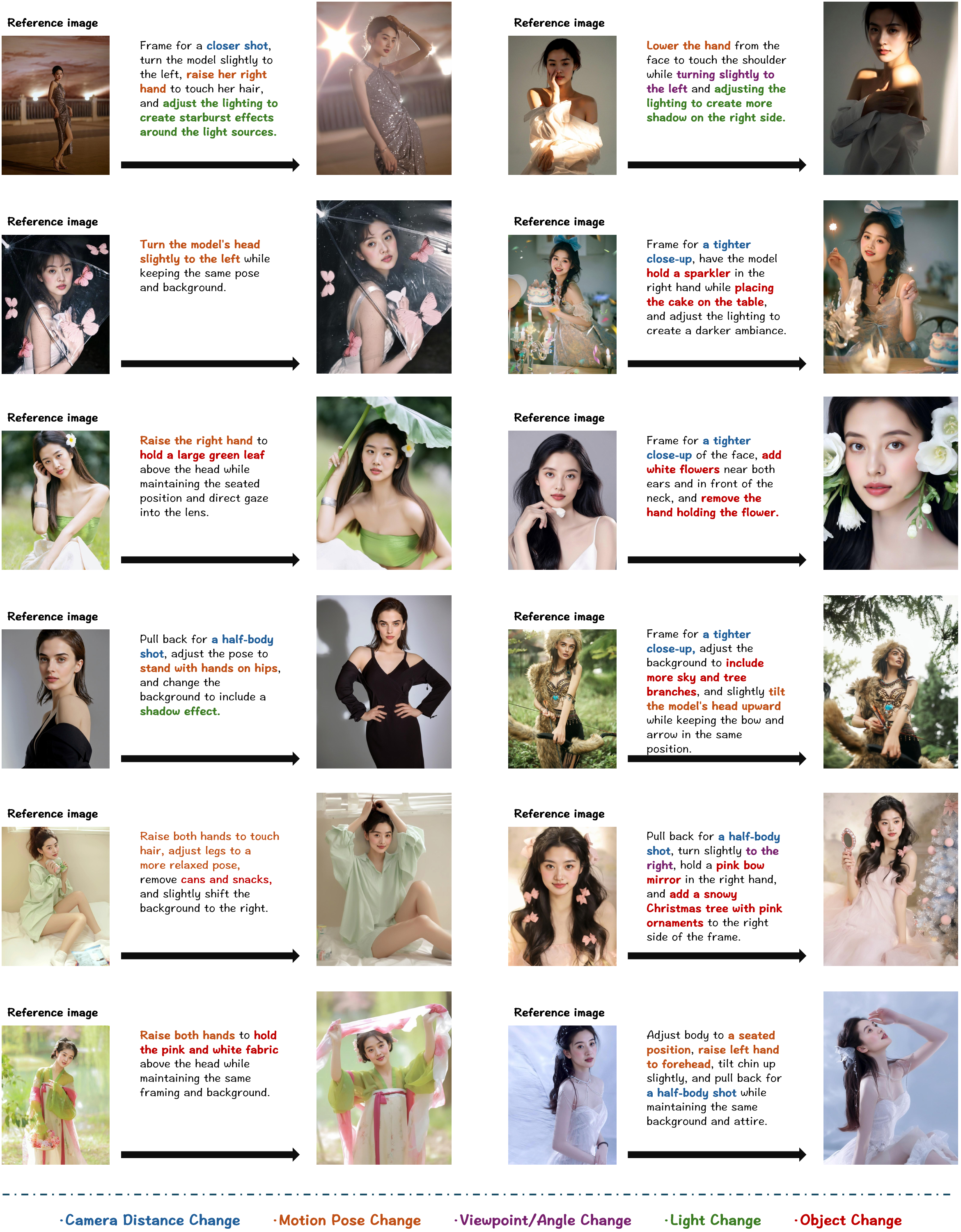} 
\vspace{-0.25in}
\caption{\textbf{Qualitative Examples of Diverse Editing Challenges.} SCheese is capable of camera distance, pose variation, and viewpoint transformation, light change and object change while preserving fine-grained details. }
\vspace{-0.2in}
\label{model_ability}
\end{figure*}

We demonstrate SCheese's capability to handle several key editing challenges in Figure~\ref{model_ability}. \textbf{Camera distance variation} includes pulling back for ``\textit{half-body shots}" or framing ``\textit{tight close-ups}", requiring scale adaptation while preserving facial details and accessories. \textbf{Pose transformation} encompasses complex body movements, such as \textit{raising hand} or adjusting to \textit{seated positions with specific hand placements}. \textbf{Viewpoint transformation} involves rotating the subject to show different profiles (e.g., ``\textit{left/right}"), requiring consistent identity preservation across viewing angles. \textbf{Light change.} includes \textit{"starburst effects"} or \textit{"more shadow"}. \textbf{Object change.} includes \textit{"adding fabric"} or \textit{"pink bow mirror"}.
As shown in Figure~\ref{model_ability}, SCheese successfully handles all these types of modifications while preserving fine-grained details from the reference image, demonstrating its effectiveness in addressing the diverse editing requirements of PCG.

\section{Additional Qualitative Results}
\begin{figure*}[t!]
\centering  
\includegraphics[width=0.9\linewidth]{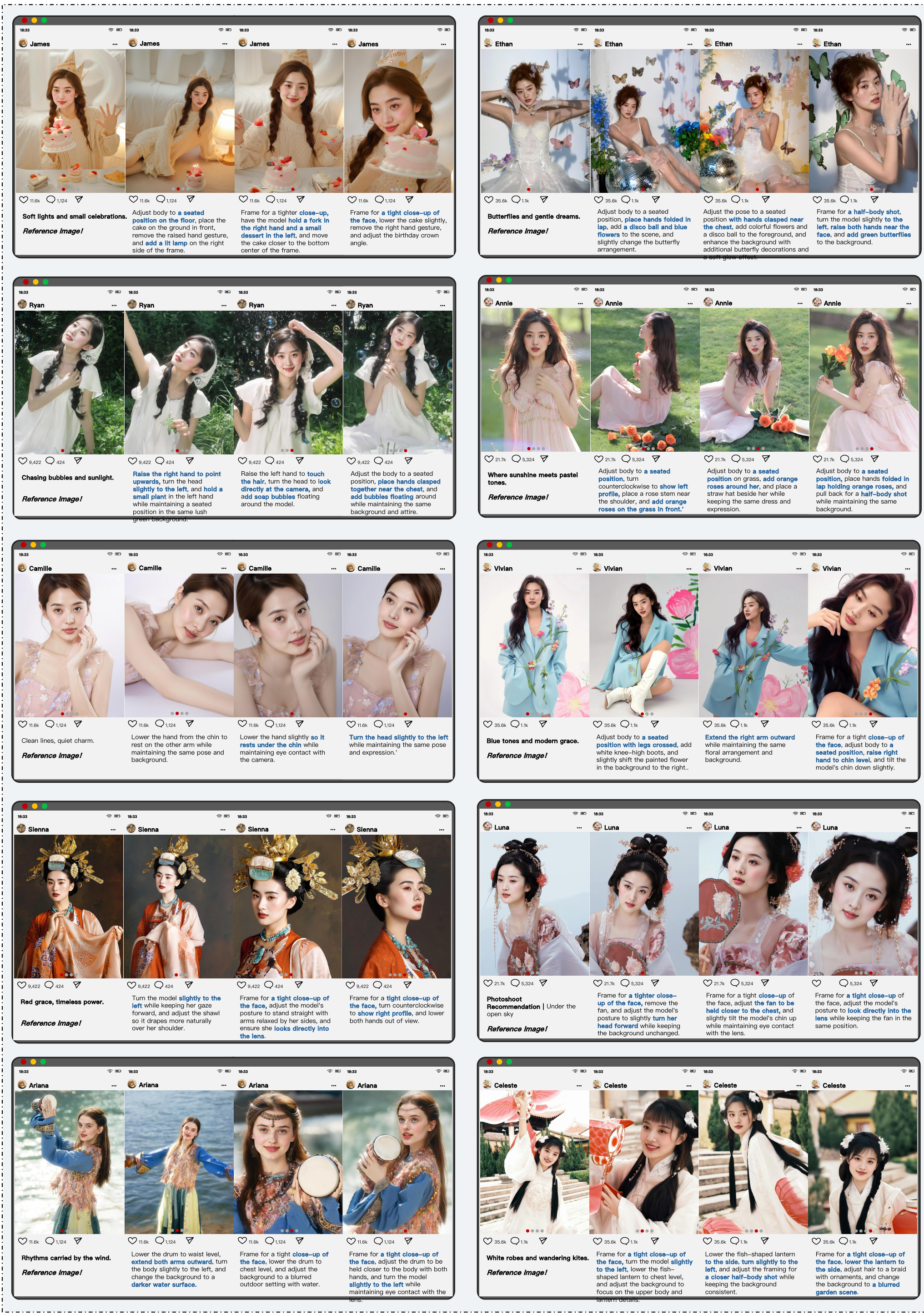} 
\vspace{-0.1in}
\caption{\textbf{More Qualitative Results.} SCheese can generate complete portrait collections with consistent identity and rich content.  }
\vspace{-0.2in}
\label{more_result}
\end{figure*}

More qualitative results are provided in Figure~\ref{more_result}. As illustrated in the figure, SCheese enables users to generate complete portrait collections with consistent identity and rich content from a single reference portrait image and multiple modification texts, demonstrating the practical utility of our approach.

\section{Data Privacy and Ethical Considerations}

To ensure ethical use and protect privacy, we implement comprehensive data anonymization procedures. First, during portrait collection selection, we exclude albums containing minors, pregnant individuals, and religious elements.To protect user privacy, we perform face replacement with synthetic virtual faces on all images. 

\textbf{Dataset Usage Restrictions.} The CHEESE dataset is intended solely for research purposes. Users must agree not to use the dataset for commercial purposes, identity recognition, or any activities that could harm individuals. The dataset is provided under strict usage terms that prohibit attempts to reverse-engineer or recover original identities.

\textbf{Data Security.} During dataset construction, all original images are processed in secure environments and permanently deleted after anonymization. Only anonymized versions are retained in the final dataset.

\end{document}